\RequirePackage{fix-cm}
\documentclass[10pt,twocolumn,letterpaper]{article}
\usepackage[T1]{fontenc}
\usepackage{times}
\usepackage{courier}
\usepackage[letterpaper,hmargin=0.65in,vmargin=0.75in]{geometry}
\usepackage{cite}
\usepackage{amsmath,amssymb,amsfonts}
\usepackage{graphicx}
\usepackage{textcomp}
\usepackage{hyperref}
\hypersetup{
  hidelinks,
  pdftitle={EndoWorld-JEPA: Motion-Guided Global--Local Latent World Modeling for Endoscopic Video Representation Learning},
  pdfauthor={Jie Wang, Shilong Yang, Tao Zhou, Juan Yu, Rong Qu, Jianfeng Ren, Zhicheng Zhang, Linlin Shen},
  pdfsubject={Endoscopic video representation learning},
  pdfkeywords={Endoscopy, JEPA, latent world model, optical flow, self-supervised learning}
}
\newcommand{\method}{EndoWorld-JEPA}
\makeatletter
\renewcommand{\subsection}{\@startsection{subsection}{2}{\z@}%
  {-3.25ex\@plus -1ex\@minus -.2ex}%
  {1.5ex\@plus .2ex}%
  {\normalfont\fontsize{11}{13.2}\selectfont\bfseries\raggedright
   \hyphenpenalty=10000\exhyphenpenalty=10000}}
\makeatother

\title{EndoWorld-JEPA: Motion-Guided Global--Local Latent World Modeling for Endoscopic Video Representation Learning}
\author{\begin{minipage}{\dimexpr\textwidth-2\tabcolsep\relax}
\centering\small
\mbox{\mbox{Jie Wang\textsuperscript{1}}, \mbox{Shilong Yang\textsuperscript{2}}, \mbox{Tao Zhou\textsuperscript{3}}, \mbox{Juan Yu\textsuperscript{4}}, \mbox{Rong Qu\textsuperscript{1}}, \mbox{Jianfeng Ren\textsuperscript{1}}, \mbox{Zhicheng Zhang\textsuperscript{5}}, and \mbox{Linlin Shen\textsuperscript{1,2}}}\\[0.75ex]
\resizebox{\linewidth}{!}{\footnotesize \textsuperscript{1}University of Nottingham\quad \textsuperscript{2}Shenzhen University\quad \textsuperscript{3}Tsinghua University\quad \textsuperscript{4}The First Affiliated Hospital of Shenzhen University\quad \textsuperscript{5}JancsiLab}
\end{minipage}}
\date{}

\begin{document}
\maketitle

\begin{abstract}
  Endoscopic video contains state transitions induced by camera displacement, deformable tissue, and instrument interaction, yet existing pre-training methods largely treat these observed dynamics implicitly. Inspired by action-conditioned joint-embedding predictive world models and by the observation that endoscopic optical flow contains both shared image-plane displacement and localized residual motion, we reformulate passive endoscopic video pre-training as causal motion-guided latent-state prediction. We present \method, to our knowledge the first latent world-model pre-training framework designed for general-purpose endoscopic video representation learning. In place of recorded actions, calibrated poses, or robot kinematics, the model conditions prediction of the next latent state on motion observed up to the current state. Its Optical Motion-guided module separates dominant displacement from residual motion and converts the resulting evidence into an ordered temporal motion condition and spatial local-motion weights. Complementary global and motion-local predictors use these outputs: the temporal condition modulates both latent transitions, whereas the local weights define an auxiliary motion-local state that extends prediction from scene-level evolution to spatially selective dynamics. Patch-wise motion weighting emphasizes localized change while retaining anatomical context, and past-motion-window dropout randomly withholds the complete motion condition to reduce dependence on unreliable motion. This global--local formulation learns predictable visual--motion dynamics in representation space without reconstructing stochastic appearance artifacts. Evaluation shows that a common pre-trained checkpoint achieves consistently strong transfer performance across four downstream benchmarks, supporting motion-guided global--local latent prediction as an effective pre-training paradigm for endoscopic video.
\end{abstract}

\par\smallskip\noindent\textbf{Keywords:}
  Endoscopy, foundation model, joint-embedding predictive architecture, latent world model, optical flow, self-supervised learning.
\par\medskip

\section{Introduction}
Endoscopy is a dynamic imaging process in which the camera continuously changes viewpoint while soft tissue deforms, instruments interact with the scene, and illumination varies with camera--surface geometry. These factors can substantially alter appearance even when anatomy or lesion identity remains unchanged, and these dynamics are sufficiently pronounced that endoscope egomotion and deformable tissue motion are themselves explicit targets of estimation and reconstruction~\cite{liu2013opticalflow,liu2025endogaussian}. Conversely, local temporal changes may reveal lesion boundaries, tissue deformation, or procedural events that are ambiguous from appearance alone. Because clinically relevant endoscopic tasks arise within dynamic examinations---even when annotations or evaluations are frame-wise---representation learning from endoscopic video should capture not only what is visible, but also how the observed state evolves.

\begin{figure}[!t]
  \centering
  \includegraphics[width=\columnwidth]{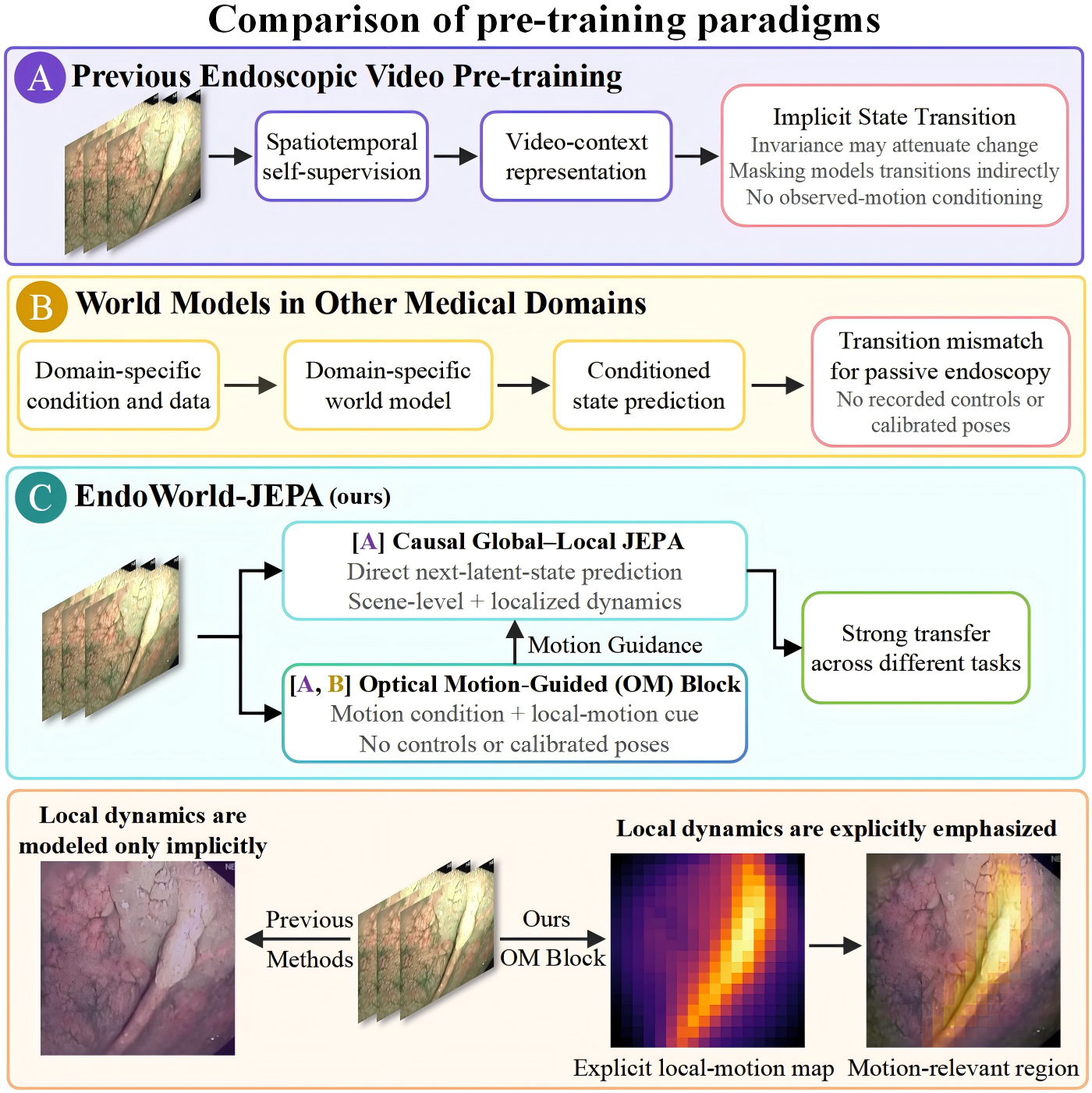}
  \caption{Comparison of three pre-training paradigms and their treatment of local dynamics. (A) Existing endoscopic video pre-training captures temporal context, but transitions remain implicit: invariance may attenuate change, masking models transitions indirectly, and observed motion does not condition prediction. (B) Medical world models rely on domain-specific data and transition conditions unavailable in passive endoscopy. (C) \method{} couples an Optical Motion-guided (OM) block that derives motion guidance from unlabeled passive video with a causal global--local JEPA, requiring no recorded controls or calibrated poses. Bottom: a schematic comparison illustrates that existing video pre-training models localized dynamics only implicitly within global latent features, whereas the OM block produces an explicit local-motion map whose patch weighting emphasizes motion-relevant regions.}
  \label{fig:pretraining_paradigms}
\end{figure}

Endoscopic foundation models have mainly focused on spatiotemporal invariance and temporal context. Endo-FM learns transferable representations from diverse endoscopic videos~\cite{wang2023endofm}, and EndoFM-LV extends masked representation learning to minute-level sequences~\cite{wang2025endofmlv}. By emphasizing content that persists across time or can be recovered from surrounding observations, these approaches learn robust temporal representations. However, invariance-oriented objectives may attenuate informative change, whereas masked reconstruction models temporal transitions only indirectly; neither explicitly conditions prediction of the next latent state on preceding image-plane motion.

Latent world modeling provides a structured alternative in which observations are encoded as states and future states are predicted under an explicit transition condition. Joint-embedding predictive architectures (JEPAs) perform this prediction in representation space rather than reconstructing pixels~\cite{assran2023ijepa,bardes2024vjepa}. This property is well matched to endoscopy: a useful representation should preserve anatomy, lesion identity, and motion-relevant structure without being required to synthesize reflections, smoke, fluid, blur, or illumination. LeWorldModel (LeWM) enables stable end-to-end next-latent-state prediction through a regularized objective~\cite{maes2026lewm}. Its conventional action-conditioned formulation, however, assumes robot commands, probe controls, or calibrated poses that are unavailable in most routine endoscopic recordings. Consequently, passive endoscopic recordings do not directly provide the transition variable assumed by standard state--action world models.

Passive video nevertheless offers observable motion evidence that can condition state transitions without physical controls. Dense optical flow requires only neighboring frames and captures both broad image displacement, predominantly associated with endoscope motion, and localized departures associated with tissue deformation or instruments. The former summarizes whole-view state evolution, whereas the latter identifies spatial regions whose dynamic evidence may be diluted by a global frame representation. Directly using raw flow magnitude as a single condition would entangle these roles and amplify errors caused by non-Lambertian surfaces, specularities, occlusion, or rapid motion. Dominant--residual decomposition allows the same motion evidence to support two complementary representations: an ordered temporal descriptor that jointly summarizes dominant displacement, flow dispersion, and residual statistics provides an observation-derived condition for latent-state transitions, while the residual-magnitude map serves as a spatial cue for locally dynamic content. Optical motion therefore supplies a transition condition and a local relevance cue, whereas JEPA learns their predictable consequences directly in representation space without reconstructing every photometric fluctuation.

Motivated by this complementarity, we propose \method, an endoscopy-oriented extension of LeWM for motion-conditioned latent-state prediction from passive recordings under a causal temporal ordering. Here, \emph{causal} refers to input availability: each prediction uses only the current visual state and motion observed before the target state. Its Optical Motion-guided (OM) module converts observed flow into a temporal motion condition and a spatial local-motion magnitude map. The temporal condition parameterizes both global and motion-local latent transitions, whereas the spatial map defines an auxiliary motion-local state, enabling complementary predictors to model scene-level evolution together with spatially selective change. Patch-wise mixed weighting retains anatomical context, and past-motion-window dropout reduces reliance on unreliable flow. We pre-train a single checkpoint on unlabeled endoscopic video and evaluate its transfer on four downstream tasks---video classification, polyp segmentation, polyp detection, and surgical phase recognition---using the corresponding PolypDiag, CVC-12k, KUMC, and Cholec80 benchmarks, which respectively assess appearance recognition, dense delineation, object localization, and procedural progression. We further define a dependency-aware cross-task ablation protocol that separates the individual and joint contributions of the proposed components. The main contributions are:
\begin{itemize}
  \item We formulate passive endoscopic video pre-training as motion-guided latent world modeling under a causal temporal ordering, using motion observed up to the current state to predict the next latent state in representation space.
  \item We introduce an Optical Motion-guided module that decomposes dense flow into ordered temporal descriptors---combining dominant displacement, flow dispersion, and residual-motion statistics---that form an observation-derived temporal condition, together with a residual local-motion magnitude map, providing transition conditioning and spatial guidance without recorded controls or calibrated poses.
  \item We introduce complementary global and motion-local JEPA objectives, together with motion--uniform mixed pooling in patch-wise motion weighting and past-motion-window dropout, to preserve scene context, emphasize localized dynamics, and reduce sensitivity to flow uncertainty.
\end{itemize}

\section{Related Work}
\subsection{Medical and Endoscopic Foundation Models}
Deep learning has advanced medical image analysis across diagnosis, prognosis, and segmentation~\cite{wang2025endocytoscopy,jiang2023biologyguided,zhang2026virtualmri}. Building on these task-specific advances, medical foundation models use large-scale pre-training to acquire representations that can be adapted across tasks and datasets. Modality-specific approaches further incorporate structural priors into their objectives. EchoFM combines consistent spatiotemporal masking with periodic contrastive learning for echocardiography~\cite{kim2025echofm}. CheXFound integrates global image features with multi-scale local radiographic features, and DCXFM combines text, image-level labels, and segmentation masks to support dense prediction~\cite{yang2025chexfound,wang2025dcxfm}. In endoscopy, Endo-FM learns spatiotemporal invariance from large video collections~\cite{wang2023endofm}, whereas EndoFM-LV uses minute-level sequences to model long-range examination context~\cite{wang2025endofmlv}. EndoDINO scales image-level self-distillation over curated frames sampled from endoscopic videos and transfers frozen encoders across recognition, segmentation, and disease-scoring tasks~\cite{dermyer2025endodino}. GastroNet-5M establishes large-scale multicenter domain-specific pre-training and evaluates its accuracy, data efficiency, and robustness across a broad range of gastrointestinal applications~\cite{jong2026gastronet5m}. EndoMamba combines bidirectional spatial modeling, causal temporal state-space modeling, and hierarchical reconstruction--alignment supervision for efficient online video inference~\cite{tian2025endomamba}. \method{} instead conditions prediction of the next latent state on previously observed short-range motion and uses the same motion evidence for spatially selective prediction.

\subsection{JEPA and Medical World Modeling}
I-JEPA learns image representations by predicting embeddings of masked regions, avoiding pixel-level reconstruction~\cite{assran2023ijepa}. V-JEPA extends feature prediction to video, and V-JEPA~2 scales action-free video pre-training before post-training an action-conditioned predictor for robotic planning~\cite{bardes2024vjepa,assran2025vjepa2}. LeWM formulates stable end-to-end next-latent-state prediction with a Gaussian latent regularizer~\cite{maes2026lewm}. Medical world models define states and transitions according to domain-specific structure. CheXWorld models local anatomy, global layout, and radiographic domain transformations~\cite{yue2025chexworld}. Medical World Model predicts disease-state changes under treatment policies~\cite{yang2025medicalworld}, while Cardiac Copilot learns probe-conditioned transitions between echocardiographic planes~\cite{jiang2024cardiaccopilot}. \method{} focuses on one-step latent-state prediction for representation learning from passive endoscopy, deriving an interpretable transition condition from preceding image-plane motion rather than recorded control signals.
Figure~\ref{fig:pretraining_paradigms} contrasts this formulation with video-context endoscopic pre-training and with medical world models whose transitions are defined by modality- or task-specific signals.

\subsection{Motion in Endoscopic Video}
Endoscopic motion couples camera movement, tissue deformation, respiration, and instrument interaction. Optical flow has been used to estimate colonoscope egomotion from live video~\cite{liu2013opticalflow} and to support camera-pose estimation~\cite{hayoz2023pose}. Video-derived geometric supervision also enables single-frame endoscopic depth estimation~\cite{liu2020endodepth}, while foundation-model-guided Gaussian splatting reconstructs deformable tissue from endoscopic sequences~\cite{liu2025endogaussian}. These methods use motion or geometry for navigation, depth estimation, or reconstruction. In contrast, \method{} uses dense flow as weak supervision for latent representation learning by separating dominant image displacement from localized residual motion.

\begin{figure*}[!t]
  \centering
  \includegraphics[width=0.80\textwidth]{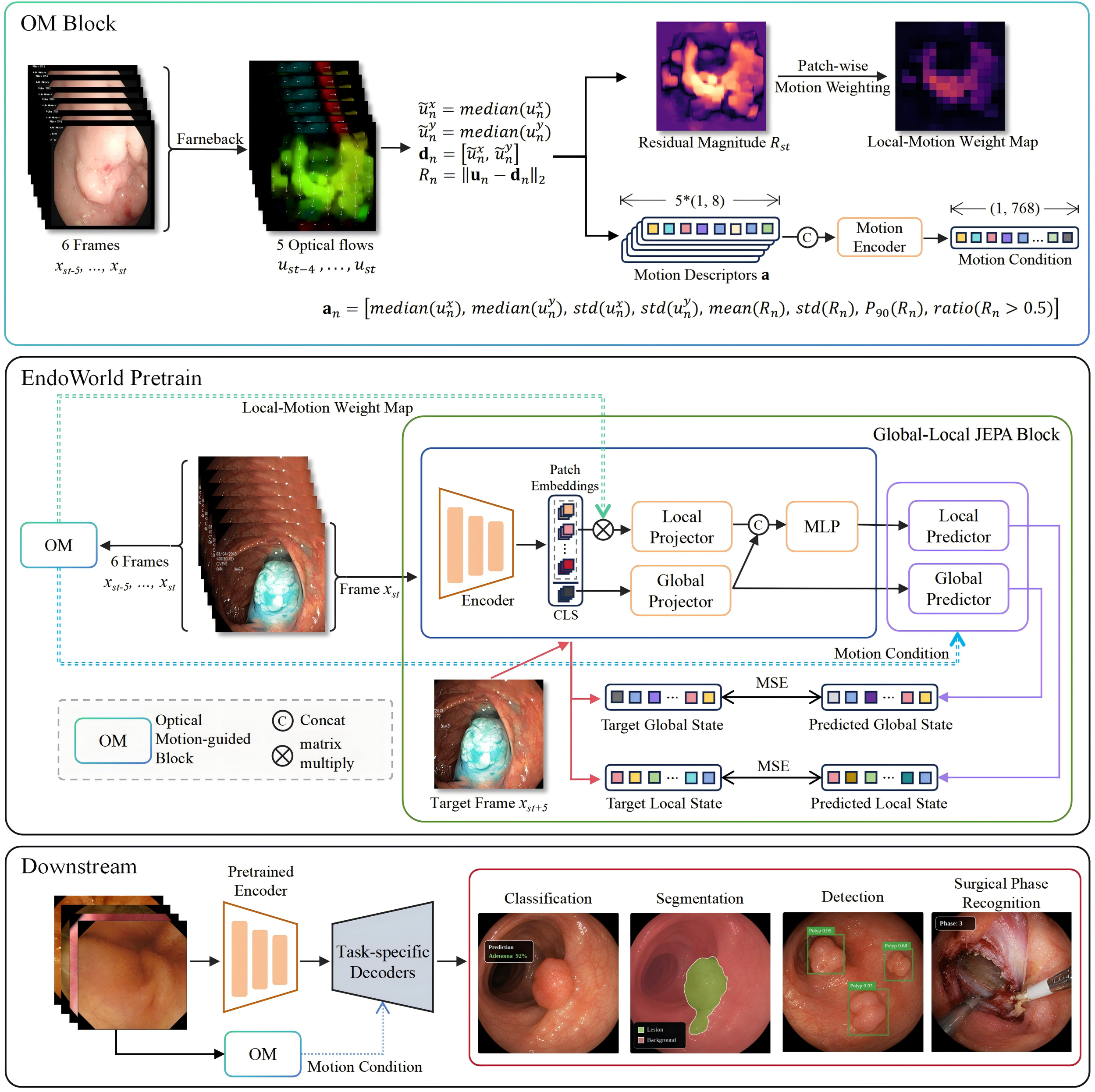}
  \iffalse

  \fi
  \caption{Overview of \method, a causal motion-conditioned JEPA world model. Top: for the implemented $K=5$ horizon, the OM block converts six consecutive observations ending at $x_{s_t}$ into five incoming optical flows, motion descriptors, a local-motion weight map, and a motion condition. Middle: the global--local JEPA block uses this five-transition motion history with the current state $x_{s_t}$ to predict local and global latent targets at $x_{s_t+5}$. Bottom: the pretrained encoder and OM block are transferred with task-specific decoders for four downstream tasks.}
  \label{fig:method}
\end{figure*}

\section{Method}
The proposed framework comprises three stages (Fig.~\ref{fig:method}): OM feature construction, global--local latent pre-training, and downstream transfer. OM derives a temporal condition and a local-motion magnitude map; patch-wise motion weighting maps the latter to weights for local-state construction. A shared visual encoder forms the current global and motion-local states, and two conditioned predictors estimate their next states. Predictor inputs are restricted to the current visual state and preceding motion; the next sampled frame supplies only latent targets. Downstream models retain the visual and motion encoders, whereas the fixed spatial weighting pathway is pre-training-only.

\subsection{World-Model Pre-Training Formulation}
Given an unlabeled endoscopic sequence $\{x_n\}_{n=0}^{T-1}$ of $T$ decoded observations, let $n$ index their order and let $t$ index sampled latent states. We deliberately use a shared observation horizon $K$ for motion aggregation and latent-state prediction. The $K$ incoming adjacent-observation transitions ending at $x_{s_t}$ form the causal motion window, while the prediction target is $x_{s_t+K}$; equivalently, $s_{t+1}=s_t+K$. Thus, motion observed over the preceding $K$ transitions conditions prediction of the latent state $K$ observation steps ahead, aligning the temporal support of the observation-derived motion condition with the prediction interval. With $s_0$ denoting the first sampled observation index, state $t$ therefore corresponds to frame $x_{s_t}$, where $s_t=s_0+tK$. During self-supervised optimization, the effective motion input is $\mathbf{A}^{\mathrm{in}}_t=\delta\mathbf{A}_t$, where $\delta\sim\mathrm{Bernoulli}(0.7)$ is sampled independently for each training pair; during evaluation and downstream adaptation, $\mathbf{A}^{\mathrm{in}}_t=\mathbf{A}_t$. \method{} predicts the next sampled latent state from the current global and motion-local states together with the effective causal motion input available at that state:
\begin{equation}
  (\hat{\mathbf{g}}_{t+1},\hat{\mathbf{l}}_{t+1}) =
  \mathcal{W}(\mathbf{g}_t,\mathbf{l}_t,\mathbf{A}^{\mathrm{in}}_t),
\end{equation}
where $\mathbf{g}_t$ and $\mathbf{l}_t$ are the global and motion-local latent states of $x_{s_t}$, $\mathbf{A}_t$ contains the $K$ most recent standardized motion descriptors ending at $x_{s_t}$, and $\mathcal{W}$ is the latent world model. Hats denote predicted states. A zero value of $\delta$ withholds the complete standardized motion window while preserving the visual input. All predictor inputs are observable by $s_t$; $x_{s_{t+1}}$ is used only as an embedding-space target.

This formulation separates visual latent-state encoding from conditional next-latent-state prediction in representation space. The motion history $\mathbf{A}_t$ is estimated from transitions preceding the current state and parameterizes the transition condition for passive recordings. The global and motion-local predictors are optimized directly in representation space.

The model contains a frame-wise visual encoder $\mathcal{E}_{\mathrm{V}}$, the OM module with learned motion encoder $\mathcal{E}_{\mathrm{M}}$, a global predictor $\mathcal{D}_{\mathrm{G}}$, and a motion-local predictor $\mathcal{D}_{\mathrm{L}}$. For each sampled state,
\begin{equation}
  \mathcal{E}_{\mathrm{V}}(x_{s_t})=
  [\mathbf{v}^{\mathrm{cls}}_t,
    \mathbf{v}^{\mathrm{patch}}_{t,1},\ldots,
  \mathbf{v}^{\mathrm{patch}}_{t,N}],
\end{equation}
where $\mathbf{v}^{\mathrm{cls}}_t$ is the CLS token, $\mathbf{v}^{\mathrm{patch}}_{t,i}$ is patch token $i$, and $N$ is the number of patch tokens. The global state is $\mathbf{g}_t=P_{\mathrm{G}}^{\mathrm{enc}}(\mathbf{v}^{\mathrm{cls}}_t)$, where $P_{\mathrm{G}}^{\mathrm{enc}}$ is the global projection MLP.

The shared encoder independently processes the current and target frames. Only the current embedding is passed to the predictors; the target embedding and target-side local-motion magnitude map $\mathbf{M}_{t+1}$ define the supervision. The encoder- and predictor-side projection MLPs use per-sample LayerNorm to avoid cross-sample statistics and map their outputs into a common loss space. The visual encoder $\mathcal{E}_{\mathrm{V}}$ is optimized end to end without a frozen or exponential-moving-average teacher. Downstream tasks retain $\mathcal{E}_{\mathrm{V}}$ and $\mathcal{E}_{\mathrm{M}}$; the global predictor is reused only by the video-classification adapter, whereas the motion-local predictor is discarded.

\subsection{Optical Motion-guided (OM) Module}
The OM module transforms optical flow observed up to the current state into a temporal condition $\mathbf{c}_t$ and a spatial local-motion magnitude map $\mathbf{M}_t$. Farneback estimation, dominant--residual decomposition, descriptor construction, map resampling, and patch-wise motion weighting are deterministic; only the temporal encoder $\mathcal{E}_{\mathrm{M}}$ is learned. Accordingly, the flow provides observation-derived weak supervision for image-plane dynamics.

\begin{figure*}[!t]
  \centering
  \includegraphics[width=\textwidth]{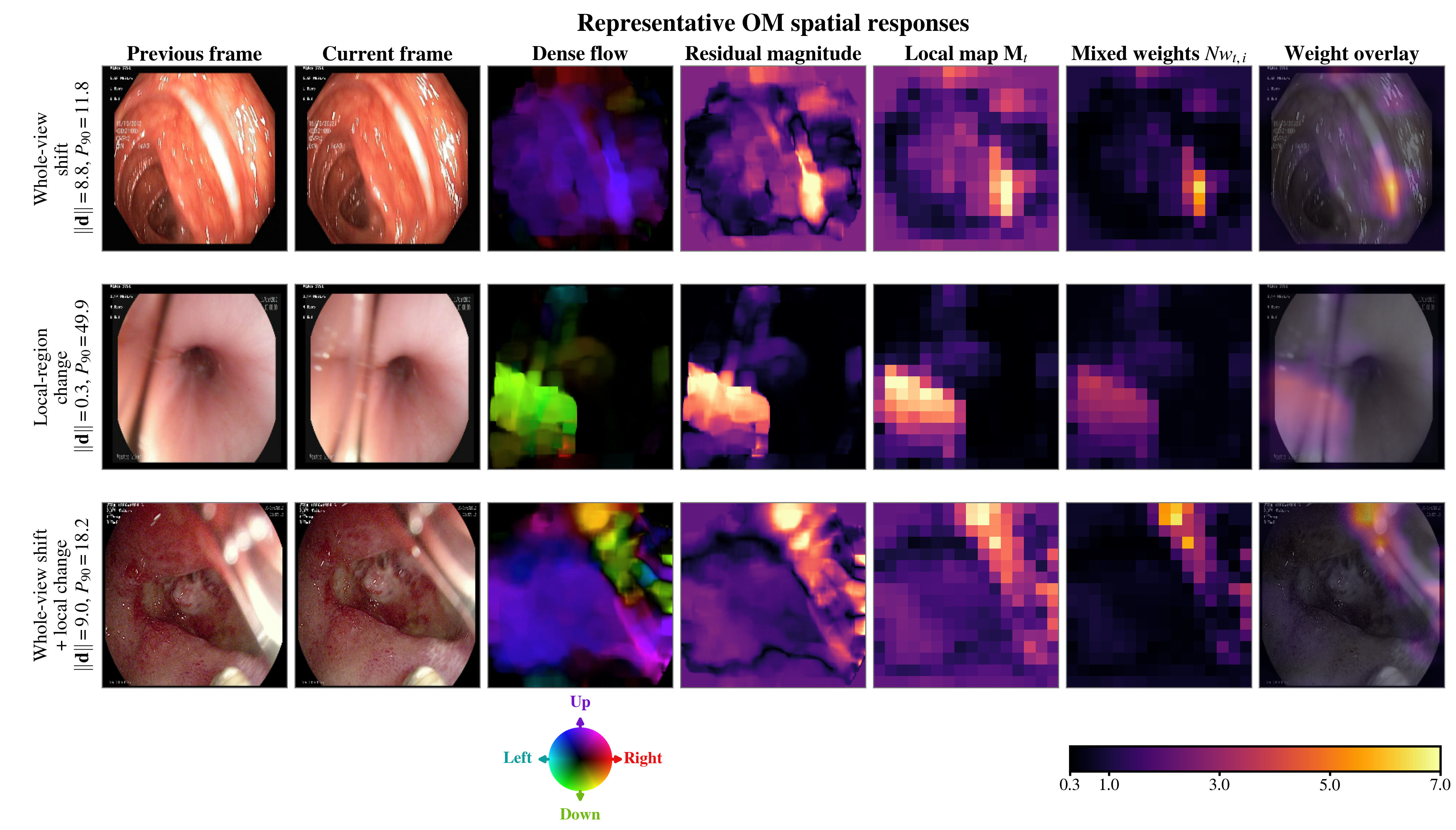}
  \caption{Representative OM responses to a whole-view shift (top), a local-region change (middle), and their combination (bottom). Columns show preceding and current frames, dense flow $\mathbf{u}_{s_t}$, residual magnitude $R_{s_t}$, local map $\mathbf{M}_t$, relative weights $Nw_{t,i}$, and the weight overlay. Dense-flow hue follows the direction key, while intensity denotes displacement magnitude. Row annotations report $\|\mathbf{d}_{s_t}\|_2$ and $P_{90}(R_{s_t})$. Residual magnitude and $\mathbf{M}_t$ share a within-row display range. Here, $Nw_{t,i}=1$ denotes uniform pooling; weight maps and overlays share the clipped $0.3$--$7.0$ scale. Display scaling does not affect OM computation.}
  \label{fig:qualitative}
\end{figure*}

For two consecutive RGB observations $x_{n-1}$ and $x_n$ with $n\geq1$, we apply field-of-view preprocessing, resize the motion input to a canonical $224\times224$ grid, and compute forward dense optical flow on this grid
\begin{equation}
  \mathbf{u}_n(\mathbf{q}) = [u^x_n(\mathbf{q}), u^y_n(\mathbf{q})],
\end{equation}
where $\mathbf{u}_n(\mathbf{q})\in\mathbb{R}^2$ maps pixel location $\mathbf{q}$ in $x_{n-1}$ toward its correspondence in $x_n$. The components $u^x_n(\mathbf{q})$ and $u^y_n(\mathbf{q})$ measure per-transition horizontal and vertical displacement in pixels on the canonical motion grid, respectively. The index $n$ associates $\mathbf{u}_n$ with the incoming transition $x_{n-1}\!\rightarrow x_n$. Farneback estimation is deterministic, computationally efficient, and requires no additional training data~\cite{farneback2003two}.

Endoscope movement often induces a broad displacement shared by much of the field of view, whereas tissue deformation, instrument interaction, and independently moving content are more localized. We approximate this dominant displacement by taking the component-wise median over image pixels, which is less sensitive than the mean to sparse extreme responses, and define the residual component as
\begin{align}
  \tilde{u}^x_n &= \mathrm{median}_{\mathbf{q}}\, u^x_n(\mathbf{q}), \\
  \tilde{u}^y_n &= \mathrm{median}_{\mathbf{q}}\, u^y_n(\mathbf{q}), \\
  \mathbf{d}_n &= [\tilde{u}^x_n,\tilde{u}^y_n], \\
  \mathbf{r}_n(\mathbf{q}) &= \mathbf{u}_n(\mathbf{q})-\mathbf{d}_n, \\
  R_n(\mathbf{q}) &= \|\mathbf{r}_n(\mathbf{q})\|_2 .
\end{align}
The component medians $\tilde{u}^x_n$ and $\tilde{u}^y_n$ form the dominant displacement $\mathbf{d}_n$; subtracting $\mathbf{d}_n$ yields the residual-flow vector $\mathbf{r}_n(\mathbf{q})$ and its Euclidean magnitude $R_n(\mathbf{q})$. The per-transition raw motion descriptor is
\begin{equation}
  \begin{aligned}
    \mathbf{a}_n = [&
      \tilde{u}^x_n,\,
      \tilde{u}^y_n,\,
      \sigma(u^x_n),\,
      \sigma(u^y_n), \\
      &
      \mu(R_n),\,
      \sigma(R_n),\,
      P_{90}(R_n),\,
    \rho_n ] ,
  \end{aligned}
\end{equation}
where $\mu(\cdot)$, $\sigma(\cdot)$, and $P_{90}(\cdot)$ are the spatial mean, standard deviation, and 90th percentile over all pixels, respectively, and $\rho_n$ is the fraction satisfying $R_n(\mathbf{q})>0.5$ pixels on the canonical motion grid. The threshold identifies spatially salient residual displacement within an observed transition. The resulting $\mathbf{a}_n$ is the raw 8-D descriptor in Fig.~\ref{fig:method}. With component-wise pre-training statistics $\boldsymbol{\mu}_{\mathrm{a}}$ and $\boldsymbol{\sigma}_{\mathrm{a}}$ computed on the same grid, its standardized form is $\mathbf{a}^{\mathrm{std}}_n=(\mathbf{a}_n-\boldsymbol{\mu}_{\mathrm{a}})/\boldsymbol{\sigma}_{\mathrm{a}}$, where the division is component-wise. At the first observation of each episode, the raw descriptor and residual map are set to zero. Missing prefix entries use the same raw-zero descriptor before standardization and therefore encode boundary padding rather than observed transitions.

The descriptor comprises two dominant-displacement, two flow-dispersion, and four residual-motion terms. For state $t$, the most recent $K$ descriptors are concatenated, using the raw-zero episode-prefix entries defined above:
\begin{equation}
  \mathbf{A}_t = [\mathbf{a}^{\mathrm{std}}_{s_t-K+1};\ldots;\mathbf{a}^{\mathrm{std}}_{s_t}],
\end{equation}
where semicolons denote ordered concatenation, giving a 40-dimensional window. Since $\mathbf{a}^{\mathrm{std}}_n$ describes $x_{n-1}\!\rightarrow x_n$, an interior window spans $K$ incoming observation transitions ending at $x_{s_t}$ and contains no target-interval motion. Both $\mathbf{A}_t$ and $\mathbf{M}_t$ are derived from apparent displacement on the canonical motion grid.

For motion-local state construction, the latest residual-magnitude field $R_{s_t}(\mathbf{q})$ is area-resampled to the ViT patch grid:
\begin{equation}
  \mathbf{M}_t =
  \mathcal{R}^{\mathrm{area}}_{16\times16}
  \left(R_{s_t}(\cdot)\right),
  \qquad m_{t,i}=[\mathbf{M}_t]_i ,
\end{equation}
where $\mathcal{R}^{\mathrm{area}}_{16\times16}$ denotes area resampling and $m_{t,i}$ is the entry at the $i$-th of $N$ flattened patch locations. The temporal window $\mathbf{A}_t$ conditions latent-state prediction, whereas $\mathbf{M}_t$ is the patch-grid local-motion magnitude map that identifies current-state patches with local dynamic evidence. We apply patch-wise motion weighting to convert its entries $m_{t,i}$ into normalized patch weights:
\begin{equation}
  \bar{m}_{t,i} = \log(1 + \max(m_{t,i},0)),
\end{equation}
\begin{equation}
  \gamma_{t,i} = \frac{\bar{m}_{t,i} - \mu(\bar{\mathbf{m}}_t)}
  {\sigma(\bar{\mathbf{m}}_t)+\epsilon},
\end{equation}
\begin{equation}
  w_{t,i} = \alpha \frac{\exp(\gamma_{t,i}/\tau)}
  {\sum_j \exp(\gamma_{t,j}/\tau)} +
  (1-\alpha)\frac{1}{N}.
\end{equation}
$\bar{m}_{t,i}$ is the nonnegative log-compressed map value, and $\bar{\mathbf{m}}_t=[\bar{m}_{t,1},\ldots,\bar{m}_{t,N}]$ collects all patch values. Their mean and standard deviation define the standardized patch score $\gamma_{t,i}$. The index $j$ ranges over all $N$ patches, $\tau$ is the softmax temperature, $\alpha$ is the motion-to-uniform mixture coefficient, and $\epsilon=10^{-6}$ ensures numerical stability. Arranged on the patch grid, $\{w_{t,i}\}_{i=1}^{N}$ forms the local-motion weight map. Mixed pooling within patch-wise motion weighting combines motion-focused softmax and uniform weights.
The motion-local state is
\begin{equation}
  \mathbf{l}_t = P_{\mathrm{L}}^{\mathrm{enc}}\left(
  \sum_{i=1}^{N} w_{t,i}\mathbf{v}^{\mathrm{patch}}_{t,i}\right),
\end{equation}

\begin{table*}[!t]
  \centering
  \caption{Comparison across four downstream endoscopic tasks under the corresponding EndoFM-LV evaluation protocols. Values are mean $\pm$ standard deviation; best results are shown in bold.}
  \label{tab:overview_comparison}
  \fontsize{8.5}{10}\selectfont
  \setlength{\tabcolsep}{4pt}
  \renewcommand{\arraystretch}{1.12}
  \begin{tabular*}{\textwidth}{@{\extracolsep{\fill}}lcccc@{}}
    \hline
    Method & \shortstack{Video Classification\\F1 (\%)} & \shortstack{Polyp Segmentation\\Dice (\%)} & \shortstack{Polyp Detection\\F1 (\%)} & \shortstack{Surgical Phase Recognition\\Accuracy (\%)} \\
    \hline
    Scratch & $83.5\pm1.3$ & $53.2\pm3.2$ & $73.5\pm4.3$ & $80.3\pm8.7$ \\
    TimeSformer~\cite{bertasius2021timesformer} & $84.2\pm0.8$ & $56.3\pm1.5$ & $75.8\pm2.1$ & $81.2\pm8.9$ \\
    CORP~\cite{hu2021corp} & $87.1\pm0.6$ & $68.4\pm1.1$ & $78.2\pm1.4$ & $82.6\pm9.6$ \\
    FAME~\cite{ding2022fame} & $85.4\pm0.8$ & $67.2\pm1.3$ & $76.9\pm1.2$ & $81.9\pm9.2$ \\
    ProViCo~\cite{park2022provico} & $86.9\pm0.5$ & $69.0\pm1.5$ & $78.6\pm1.7$ & $82.3\pm8.5$ \\
    VCL~\cite{qian2022vcl} & $87.6\pm0.6$ & $69.1\pm1.2$ & $78.1\pm1.9$ & $82.7\pm9.4$ \\
    ST-Adapter~\cite{pan2022stadapter} & $84.8\pm0.7$ & $64.3\pm1.9$ & $74.9\pm2.9$ & $81.0\pm8.7$ \\
    Endo-FM~\cite{wang2023endofm} & $90.7\pm0.4$ & $73.9\pm1.2$ & $84.1\pm1.3$ & $82.8\pm9.1$ \\
    EndoSSL~\cite{hirsch2023endossl} & $91.3\pm0.6$ & $75.1\pm1.6$ & $83.2\pm1.6$ & $83.0\pm8.0$ \\
    EndoMamba~\cite{tian2025endomamba} & $95.0\pm1.3$ & $85.4\pm0.2$ & $90.1\pm1.4$ & $84.3\pm7.3$ \\
    EndoFM-LV~\cite{wang2025endofmlv} & $96.3\pm0.7$ & $83.2\pm0.3$ & $92.5\pm1.2$ & $86.1\pm7.1$ \\
    \method{} (ours) & $\boldsymbol{98.3\pm0.3}$ & $\boldsymbol{87.2\pm0.6}$ & $\boldsymbol{95.2\pm0.9}$ & $\boldsymbol{89.7\pm6.3}$ \\
    \hline
  \end{tabular*}
\end{table*}

where $P_{\mathrm{L}}^{\mathrm{enc}}$ is the local-state projection MLP and $w_{t,i}$ is the final normalized weight assigned to patch token $\mathbf{v}^{\mathrm{patch}}_{t,i}$. The uniform component prevents the local state from concentrating entirely on isolated flow responses.

Patch-wise motion weighting limits isolated softmax peaks while retaining spatial selectivity. The temporal window $\mathbf{A}_t$ conditions both predictors, and the local-motion weight map $\{w_{t,i}\}_{i=1}^{N}$ assigns greater weight to locally dynamic patches in the current frame. The independently computed target magnitude map $\mathbf{M}_{t+1}$ is converted into $\{w_{t+1,i}\}_{i=1}^{N}$ and contributes only to the local target. Both magnitude maps characterize image-plane motion and may respond to instruments, occlusion boundaries, specularities, and flow errors in addition to tissue motion. The lower bound $(1-\alpha)/N$ on every patch weight preserves surrounding anatomical context.

Figure~\ref{fig:qualitative} illustrates OM responses to three representative image-plane motion patterns. Predominantly whole-view displacement produces a large median-flow magnitude and attenuated residuals after common-motion subtraction. Localized change produces a concentrated residual despite limited shared displacement, whereas mixed motion contains both responses. In each case, $\mathbf{M}_t$ retains residual-motion magnitude, and patch-wise motion weighting converts it into the local-motion weight map $\{w_{t,i}\}_{i=1}^{N}$ used to pool current-frame patch tokens. The overlays visualize this spatial emphasis while confirming retention of full-frame context.

\subsection{Global--Local Latent State Prediction}
\method{} learns passive endoscopic video representations through causal motion-conditioned latent-state prediction. The global latent state $\mathbf{g}_t$ represents the complete frame, whereas the motion-local latent state $\mathbf{l}_t$ pools patch tokens with the local-motion weights defined above.

Within the temporal branch of OM, the standardized past-motion window $\mathbf{A}_t$ provides the observation-derived transition signal. The motion encoder maps its effective form to the encoded motion condition:
\begin{equation}
  \mathbf{c}_t = \mathcal{E}_{\mathrm{M}}(\mathbf{A}^{\mathrm{in}}_t).
\end{equation}
The predictor receives only the current sampled state and its encoded causal motion condition. The global branch specializes the LeWM latent-state prediction formulation to
\begin{equation}
  \hat{\mathbf{g}}_{t+1} = P_{\mathrm{G}}^{\mathrm{out}}\!\left(
    \mathcal{D}_{\mathrm{G}}(
  \mathbf{g}_t,\mathbf{c}_t)\right).
\end{equation}
The motion-local branch first forms the fused state
\begin{equation}
  \mathbf{f}_t = P_{\mathrm{F}}([\mathbf{g}_t;\mathbf{l}_t]).
\end{equation}
The fusion supplies scene context through $\mathbf{g}_t$ before local prediction. The local branch then predicts
\begin{equation}
  \hat{\mathbf{l}}_{t+1} = P_{\mathrm{L}}^{\mathrm{out}}\!\left(
    \mathcal{D}_{\mathrm{L}}(
  \mathbf{f}_t,\mathbf{c}_t)\right).
\end{equation}
The operator $[\cdot;\cdot]$ denotes concatenation, $P_{\mathrm{F}}$ is the fusion MLP, and $P_{\mathrm{G}}^{\mathrm{out}}$ and $P_{\mathrm{L}}^{\mathrm{out}}$ are the corresponding predictor-output projections. The pairs $(\hat{\mathbf{g}}_{t+1},\mathbf{g}_{t+1})$ and $(\hat{\mathbf{l}}_{t+1},\mathbf{l}_{t+1})$ define global and local supervision, respectively.

Each predictor processes one current-state token and returns one next-latent-state estimate. The encoded condition $\mathbf{c}_t$ modulates every predictor block through adaptive LayerNorm with zero-initialized residual gates. The global and auxiliary local predictors have separate parameters; their dimensions and depths are specified in the implementation details.

\subsection{Training Objective and Robustness}
The past-motion-window dropout defined above retains the complete standardized motion window when $\delta=1$ and replaces its effective input with zeros when $\delta=0$. This sample-wise perturbation preserves descriptor ordering and forces the current visual state to remain predictive when the motion condition is withheld; it is confined to self-supervised optimization.

The global and local predictors are optimized jointly for the single target in each training pair. For batch size $B$ and loss-embedding dimension $d_z=768$, let $b\in\{1,\ldots,B\}$ index training pairs and let the superscript $(b)$ select the corresponding sample. Using $\|\cdot\|_2$ for the Euclidean norm, the implemented mean-squared losses are
\begin{equation}
  \mathcal{L}_{\mathrm{global}} =
  \frac{1}{Bd_z}\sum_{b=1}^{B}
  \left\| \hat{\mathbf{g}}^{(b)}_{t+1} - \mathbf{g}^{(b)}_{t+1} \right\|_2^2,
\end{equation}
\begin{equation}
  \mathcal{L}_{\mathrm{local}} =
  \frac{1}{Bd_z}\sum_{b=1}^{B}
  \left\| \hat{\mathbf{l}}^{(b)}_{t+1} - \mathbf{l}^{(b)}_{t+1} \right\|_2^2,
\end{equation}
\begin{equation}
  \mathcal{L} =
  \mathcal{L}_{\mathrm{global}}
  + \lambda_{\mathrm{L}}\mathcal{L}_{\mathrm{local}}
  + \lambda_{\mathrm{SIG}} \mathcal{L}_{\mathrm{SIGReg}}.
\end{equation}
In the total objective, $\mathcal{L}_{\mathrm{SIGReg}}$ is the distributional regularizer, and $\lambda_{\mathrm{L}}$ and $\lambda_{\mathrm{SIG}}$ weight the local prediction and SIGReg terms, respectively. SIGReg constrains the encoded global states $\mathbf{g}_t$ and $\mathbf{g}_{t+1}$ and prevents their collapse during end-to-end training. Prediction in representation space emphasizes predictable anatomical and dynamic structure without requiring exact reconstruction of reflections, smoke, blur, fluid, or illumination changes.

SIGReg projects the encoded current and target embeddings onto 1,024 randomly sampled unit directions and compares their empirical characteristic functions with that of an isotropic Gaussian at 17 knots. Applying this constraint to both encoded frames preserves dispersion without introducing negative pairs or a separate target network. The local term acts as an auxiliary spatial constraint derived from noisier residual flow, while the global objective remains the principal state-prediction signal.

\section{Experiments and Results}

\subsection{Pre-training Data and Evaluation Protocols}
\noindent\textit{Self-supervised pre-training data:}
We construct the unlabeled pre-training corpus from HyperKvasir Video~\cite{borgli2020hyperkvasir}, Kvasir-Capsule~\cite{smedsrud2021kvasircapsule}, and LDPolypVideo~\cite{ma2021ldpolypvideo}. After excluding video segments that overlap the PolypDiag validation partition, the corpus contains 6.44 million frames from 587 videos. By frame count, this corpus is smaller than those used by EndoFM-LV (over 13 million)~\cite{wang2025endofmlv} and EndoMamba (11.30 million)~\cite{tian2025endomamba}. Despite using fewer pre-training frames, the proposed model achieves stronger transfer performance across the four evaluated downstream tasks. Unlabeled segments overlapping the PolypDiag training partition are retained, and the complete filtered corpus is used for self-supervised optimization without an internal hold-out subset.

\noindent\textit{Downstream evaluation:}
We assess transfer across four task categories using the corresponding EndoFM-LV data partitions and primary metrics~\cite{wang2025endofmlv}. Visual and motion features are constructed causally at every sampled time: the frame at that time is encoded visually, OM uses at most $K$ preceding transitions, missing prefix entries are zero padded, and no subsequent observation is used.

\noindent\textit{Video classification:}
The PolypDiag split contains 173 training and 80 validation videos. Videos are sampled at 2 fps, with 16--128 retained frames per video, and performance is measured by video-level F1.

\noindent\textit{Polyp segmentation:}
The CVC-12k protocol contains 401 training and 171 evaluation frames, and performance is measured by Dice score.

\noindent\textit{Polyp detection:}
The processed KUMC train2019 and val2019 partitions contain 15,578 training and 4,254 validation frames, respectively, and performance is measured by frame-level F1.

\noindent\textit{Surgical phase recognition:}
For Cholec80, videos 1--40 are used for training and 41--80 for testing, producing 85,984 training and 97,874 test windows. Ten-frame sequences are sampled at 1 fps and terminate at the supervised target; performance is measured by video-mean accuracy.

\begin{table*}[!t]
  \centering
  \caption{Dependency-aware component ablation and motion-design controls. Values are shown as mean $\pm$ standard deviation over three runs using seeds 0, 1, and 2. G: global prediction; TC: OM temporal conditioning; ML: motion-local prediction; MP: motion--uniform mixed pooling within patch-wise motion weighting; WD: past-motion-window dropout.}
  \label{tab:component_ablation}
  \footnotesize
  \setlength{\tabcolsep}{2pt}
  \renewcommand{\arraystretch}{1.12}
  \begin{tabular*}{\textwidth}{@{\extracolsep{\fill}}lccccc cccc@{}}
    \hline
    Variant & G & TC & ML & MP & WD & \shortstack{Classification\\F1 (\%)} & \shortstack{Segmentation\\Dice (\%)} & \shortstack{Detection\\F1 (\%)} & \shortstack{Phase Recognition\\Accuracy (\%)} \\
    \hline
    Global prediction baseline & $\checkmark$ & -- & -- & -- & -- & $91.0\pm0.9$ & $73.7\pm2.0$ & $83.9\pm1.5$ & $83.0\pm6.8$ \\
    \hline
    $+$ temporal conditioning & $\checkmark$ & $\checkmark$ & -- & -- & -- & $95.0\pm0.7$ & $79.3\pm1.5$ & $89.0\pm1.4$ & $86.0\pm6.3$ \\
    $+$ temporal conditioning + dropout & $\checkmark$ & $\checkmark$ & -- & -- & $\checkmark$ & $95.7\pm0.3$ & $80.2\pm1.2$ & $89.8\pm1.2$ & $86.7\pm6.4$ \\
    \hline
    $+$ motion-local prediction & $\checkmark$ & -- & $\checkmark$ & -- & -- & $93.4\pm0.6$ & $81.1\pm1.3$ & $89.6\pm1.3$ & $85.0\pm6.7$ \\
    $+$ motion-local + mixed pooling & $\checkmark$ & -- & $\checkmark$ & $\checkmark$ & -- & $94.2\pm0.5$ & $82.7\pm1.2$ & $90.9\pm1.1$ & $85.7\pm7.0$ \\
    \hline
    $+$ temporal conditioning + motion-local & $\checkmark$ & $\checkmark$ & $\checkmark$ & -- & -- & $97.0\pm0.6$ & $84.9\pm1.2$ & $93.4\pm1.2$ & $88.1\pm7.1$ \\
    $+$ temporal conditioning + motion-local + mixed pooling & $\checkmark$ & $\checkmark$ & $\checkmark$ & $\checkmark$ & -- & $97.7\pm0.4$ & $86.1\pm0.9$ & $94.4\pm0.9$ & $88.9\pm6.3$ \\
    $+$ temporal conditioning + motion-local + dropout & $\checkmark$ & $\checkmark$ & $\checkmark$ & -- & $\checkmark$ & $97.5\pm0.3$ & $85.7\pm0.8$ & $94.1\pm1.0$ & $89.0\pm6.6$ \\
    \hline
    Full + shuffled motion condition & $\checkmark$ & $\checkmark$ & $\checkmark$ & $\checkmark$ & $\checkmark$ & $94.4\pm0.7$ & $82.4\pm1.2$ & $90.7\pm1.7$ & $85.6\pm6.5$ \\
    Dense-flow magnitude only & $\checkmark$ & -- & $\checkmark$ & $\checkmark$ & -- & $92.8\pm0.7$ & $78.9\pm1.3$ & $87.8\pm1.6$ & $84.3\pm7.7$ \\
    \hline
    Full \method{} & $\checkmark$ & $\checkmark$ & $\checkmark$ & $\checkmark$ & $\checkmark$ & $98.3\pm0.3$ & $87.2\pm0.6$ & $95.2\pm0.9$ & $89.7\pm6.3$ \\
    \hline
  \end{tabular*}
  \vspace{2pt}
  \begin{minipage}{0.98\textwidth}
    \footnotesize
    MP requires ML; WD requires TC. Pre-training shuffle reassigns intact $\mathbf{A}_t$ across episodes while $\mathbf{M}_t$ remains aligned. The dense-flow-only control omits TC and WD and defines $\mathbf{M}_t$ by area-resampling $\|\mathbf{u}_{s_t}(\cdot)\|_2$ to the patch grid. Metrics follow Table~\ref{tab:overview_comparison}.
  \end{minipage}
\end{table*}

\subsection{Implementation Details}
\noindent\textit{Pre-training preprocessing:}
Visual inputs are resized to $224\times224$. After the applicable field-of-view preprocessing, the dedicated motion input is likewise resized to the canonical $224\times224$ grid before Farneback estimation. Farneback uses pyramid scale 0.5, three levels, a 15-pixel window, three iterations, polynomial neighborhood size 5, and standard deviation 1.2. Residual magnitudes are area-resampled to the patch grid, and motion descriptors are standardized using statistics computed over the complete pre-training corpus on this canonical grid. Temporal adjacency follows decoded observation order.

\noindent\textit{Pre-training architecture and optimization:}
The model uses a ViT-Base encoder with patch size 14 and hidden dimension 768. For each sampled state, the motion encoder $\mathcal{E}_{\mathrm{M}}$ receives the five ordered eight-dimensional descriptors as a 40-dimensional vector. A $1\times1$ channel projection ($40\rightarrow10$) followed by a two-layer SiLU MLP ($10\rightarrow3072\rightarrow768$) produces a single condition token $\mathbf{c}_t\in\mathbb{R}^{1\times768}$. The global and local projection MLPs have a hidden dimension of 2,048 and use per-sample LayerNorm; the global and local conditional predictors contain six and two layers, respectively, with 12 attention heads and a 3,072-dimensional feed-forward layer. Transformer dropout is 0.2 and embedding dropout is 0.1. Following the temporal blocking setting of LeWM~\cite{maes2026lewm}, the shared observation horizon is fixed at $K=5$. We retain $\lambda_{\mathrm{SIG}}=0.09$, which lies near the optimum of LeWM's reported sensitivity study within its stable $0.01$--$0.2$ interval. Because the patch scores are standardized, $\tau=1.0$ retains their unit scale. We fix $\alpha=0.7$, $\lambda_{\mathrm{L}}=0.05$, and the past-motion-window dropout probability to 0.3 throughout all experiments; these choices allocate 30\% uniform spatial mass, keep local prediction auxiliary, and retain motion conditioning in 70\% of training pairs, respectively. Optimization runs for 200 epochs and uses AdamW with a learning rate of $3\times10^{-5}$, weight decay of $5\times10^{-3}$, batch size 64, bfloat16 precision, gradient clipping at 1.0, and a linear-warmup cosine schedule. The self-supervised objective is averaged across all corpus mini-batches within each epoch, and the epoch-end checkpoint with the lowest mean initializes all downstream tasks. Neither downstream annotations nor downstream validation metrics enter checkpoint selection.

\noindent\textit{Common downstream adaptation:}
Visual inputs are resized to $224\times224$ for PolypDiag and CVC-12k and to $640\times640$ for KUMC. Cholec80 frames undergo fixed field-of-view cropping, $250\times250$ resizing, and $224\times224$ cropping. Regardless of the task-specific visual resolution, OM operates on the canonical $224\times224$ motion grid. Within each run, all tasks initialize from the common pre-training checkpoint and fuse $\mathbf{c}_t$, but not $\mathbf{M}_t$; the local-motion magnitude map, its derived weight map, and the motion-local predictor are pre-training-only. Across all four tasks, the transferred visual and motion encoders remain frozen, and only the task-specific adapters are optimized. The video-classification adapter additionally uses descriptor-derived frame weights and is the only one to reuse the frozen global predictor. The complete adaptation protocol is repeated three times using seeds 0, 1, and 2, and Table~\ref{tab:overview_comparison} reports the mean and standard deviation across these runs.

\subsection{Video Classification}
For video-level appearance recognition, descriptor-derived magnitudes weight sampled frame features, and an MLP combines the pooled visual state, mean encoded condition, temporal statistics, and one global-predictor feature computed from a checkpoint-compatible current-state input. On the PolypDiag benchmark, \method{} achieves $98.3\pm0.3$\% F1.

\subsection{Polyp Segmentation}
Dense lesion delineation uses a hybrid decoder that combines ViT layers 3, 6, 9, and 12 with a shallow convolutional path. The condition $\mathbf{c}_t$ is projected to biases at the four decoder scales, and the decoder is trained with a weighted binary-cross-entropy--Dice loss. On the CVC-12k benchmark, the model achieves $87.2\pm0.6$\% Dice.

\subsection{Polyp Detection}
Lesion localization is performed by a compact feature pyramid applied to the target frame. The projected condition is added before heads that predict a center heatmap, offsets, and box dimensions using focal and smooth-$\ell_1$ losses. The detector obtains $95.2\pm0.9$\% F1 on the KUMC benchmark.

\subsection{Surgical Phase Recognition}
Procedural progression is modeled by adding the projected condition to spatial tokens before divided space--time decoding and concatenating it with decoded visual and elapsed-time features before a unidirectional LSTM. Causal smoothing uses no later frame. On the Cholec80 benchmark, the model achieves $89.7\pm6.3$\% video-mean accuracy.

\noindent\textit{Cross-task comparison:}
Baselines in Table~\ref{tab:overview_comparison} follow the corresponding EndoFM-LV evaluation protocols; statistics are taken from its unified evaluation~\cite{wang2025endofmlv}, the two published EndoMamba entries~\cite{tian2025endomamba}, and matched-protocol EndoMamba runs on the remaining tasks. A common \method{} checkpoint achieves the highest reported mean in every task category, exceeding the strongest previous values by 2.0, 1.8, 2.7, and 3.6 points for video classification, polyp segmentation, polyp detection, and surgical phase recognition, respectively. The consistent gains support the transferability of the learned visual--motion representation across task types.

\subsection{Component Ablation Study}
Table~\ref{tab:component_ablation} reports independently pre-trained variants with global prediction retained throughout. Each variant is independently pre-trained and evaluated in three runs using seeds 0, 1, and 2. Across rows, the visual inputs, pre-training data, normalization, optimization budgets, checkpoint selection, and downstream adaptation protocol are identical; mean and standard deviation over the three runs are reported. Temporal conditioning and motion-local prediction are tested independently and jointly; mixed pooling requires the local branch, and past-motion-window dropout requires temporal conditioning. In variants without TC, every active predictor receives $\mathbf{c}_t=\mathbf{0}$; ML variants without MP set $\alpha=1$, retaining motion-only softmax pooling. The motion-design controls are separated from the component rows. The shuffled control retains all modules but reassigns intact windows across episodes, preserving motion statistics while removing state correspondence. The dense-flow-only control removes descriptor-based temporal conditioning and area-resamples the aligned field $\|\mathbf{u}_{s_t}(\cdot)\|_2$ to obtain $\mathbf{M}_t$; consequently, TC and WD are inactive while G, ML, and MP are retained.

The global-prediction baseline performs at a comparable level to Endo-FM across the four tasks. Temporal conditioning adds 3.0--5.6 points. Motion-local prediction adds 2.0--7.4 points, with substantially larger gains of 7.4 and 5.7 points for segmentation and detection than for classification and phase recognition. Combining both components improves the baseline by 6.0, 11.2, 9.5, and 5.1 points for classification, segmentation, detection, and phase recognition, respectively. Adding mixed pooling or past-motion-window dropout to this combination contributes 0.7--1.2 and 0.5--0.9 points, respectively, while using both yields a further 1.3--2.3 points. During pre-training, shuffling affects only $\mathbf{A}_t$: the current image and aligned $\mathbf{M}_t$ remain intact, so the predictor can attenuate the mismatch and remain close to the motion-local plus mixed-pooling variant. The consistent 3.9--4.8-point reduction under shuffled conditioning indicates that temporal guidance relies on correct state--motion correspondence rather than on motion-descriptor statistics alone. Under the matched G+ML+MP setting, direct dense-flow magnitude trails residual-motion weighting by 1.4, 3.8, 3.1, and 1.4 points, respectively. The larger reductions for segmentation and detection support dominant-motion removal before spatial guidance.

The global and local objectives impose complementary inductive biases. Global prediction preserves anatomy, viewpoint, and scene organization, whereas motion-local prediction emphasizes sparse changes that may be attenuated by global pooling. Consistent with this distinction, temporal conditioning improves all four tasks, and the local branch provides its largest gains for segmentation and detection. Median-flow subtraction establishes a robust image-plane reference, although residual responses can also arise from occlusion boundaries, specularities, instruments, or flow errors. Within patch-wise motion weighting, mixed pooling limits domination by a small number of high-response patches, and past-motion-window dropout maintains the predictive contribution of visual semantics when motion is unreliable. The fixed spatial weighting pathway is excluded from downstream heads but shapes the transferred visual encoder through the auxiliary pre-training objective. By comparison, the learned condition $\mathbf{c}_t$ is fused at the representation level appropriate to each task, and the global predictor is reused only by the video-classification adapter. This transfer strategy preserves a common representation while allowing task-specific adaptation.

\section{Discussion}
Across four endoscopic tasks, motion-conditioned latent-state prediction provides a transferable pre-training signal for classification, segmentation, detection, and surgical phase recognition. Using a common checkpoint with task-specific heads, the results provide preliminary evidence that observation-indexed OM can transfer across datasets with heterogeneous temporal sampling. This pattern is consistent with cadence diversity acting as implicit temporal augmentation, while prospective evaluation across acquisition rates remains necessary.

The use of a common pre-trained checkpoint across four task-specific heads supports broad transferability, although final performance also depends on each downstream adapter. In contrast to objectives that enforce temporal invariance or reconstruct appearance details, the JEPA objective emphasizes content with predictable representation-level evolution. OM further reduces prediction ambiguity by conditioning the predictors on motion observed only up to the current state, thereby preserving causal feature construction in passive recordings. Optical flow and motion descriptors are precomputed offline during self-supervised pre-training. At downstream inference, OM can be updated causally with one Farneback estimate and lightweight descriptor encoding per new observation, without introducing an additional learned optical-flow network.

Several limitations remain. Farneback flow is efficient and deterministic, but degrades under rapid motion, large occlusion, weak texture, smoke, and severe illumination change. Component-wise median subtraction captures a dominant translation-like displacement but cannot fully factor rotation, zoom, depth-dependent parallax, or multiple independently moving regions. The fixed $K$-transition condition represents one observation scale; long-horizon stability, uncertainty calibration, counterfactual control, and planning remain outside this one-step formulation. Control-aware forecasting would additionally require physical control signals. The frame-wise visual encoder also provides a path to adaptation without temporal inputs when only isolated images are available; dedicated still-image validation remains future work. The external comparisons include different task-specific pipelines, so final performance reflects adapter design as well as pre-training quality. Although all our protocols use the same checkpoint and causal alignment, broader conclusions require prospective evaluation across institutions, devices, operators, and acquisition conditions.

Future work should explore confidence-aware flow, controlled temporal resampling, adaptive prediction horizons, and synchronized kinematics to test motion reliability and cadence robustness and to compare observation-derived conditions with physical controls.

\section{Conclusion}
We introduced \method, a causal motion-conditioned JEPA world model for passive endoscopic video. OM converts recent dense flow into an observation-derived temporal motion condition and local-motion cue without recorded controls or calibrated poses. Dominant--residual decomposition and complementary global--local prediction preserve scene context while emphasizing localized dynamics; patch-wise mixed weighting and past-motion-window dropout improve robustness, and latent prediction avoids reconstructing stochastic appearance artifacts. A common pre-trained visual--motion checkpoint achieves the highest reported mean across four downstream task categories. Ablations link these gains to correct state--motion alignment and residual guidance, supporting causal motion-conditioned latent-state prediction as an effective and transferable pre-training principle for endoscopy.

\begingroup
\small
\bibliographystyle{IEEEtran}
\bibliography{references}
\endgroup

\end{document}